\documentclass[10pt,conference]{IEEEtran}
\IEEEoverridecommandlockouts
\usepackage{cite}
\usepackage{amsmath,amssymb,amsfonts}
\usepackage{algorithmic}
\usepackage{graphicx}
\usepackage{textcomp}
\usepackage{xcolor}
\def\BibTeX{{\rm B\kern-.05em{\sc i\kern-.025em b}\kern-.08em
    T\kern-.1667em\lower.7ex\hbox{E}\kern-.125emX}}
\begin{document}
  
\title{Risk-Calibrated Learning:\\
Minimizing Fatal Errors in Medical AI
}

\author{\IEEEauthorblockN{Abolfazl Mohammadi-Seif}
\IEEEauthorblockA{\textit{Department of Engineering} \\
\textit{Universitat Pompeu Fabra}\\
Barcelona, Spain \\
abolfazl.mohammadiseif@upf.edu}
\and
\IEEEauthorblockN{Ricardo Baeza-Yates}
\IEEEauthorblockA{\textit{Department of Engineering} \\
\textit{Universitat Pompeu Fabra}\\
Barcelona, Spain \\
rbaeza@ieee.org}
\thanks{This work has been accepted for publication in the Proceedings of the 2026 International Joint Conference on Neural Networks (IJCNN 2026). The final published version should be cited.}
}

\maketitle

\begin{abstract}
Deep learning models often achieve expert-level accuracy in medical image classification but suffer from a critical flaw: \textit{semantic incoherence.} These high-confidence mistakes that are semantically incoherent (e.g., classifying a malignant tumor as benign) fundamentally differ from acceptable errors which stem from visual ambiguity. Unlike safe, fine-grained disagreements, these fatal failures erode clinical trust. To address this, we propose Risk-Calibrated Learning, a technique that explicitly distinguishes between visual ambiguity (fine-grained errors) and catastrophic structural errors. By embedding a confusion-aware clinical severity matrix $\mathbf{M}$ into the optimization landscape, our method suppresses critical errors (false negatives) without requiring complex architectural changes. We validate our approach in four different imaging modalities: Brain Tumor MRI, ISIC 2018 (Dermoscopy), BreaKHis (Breast Histopathology), and SICAPv2 (Prostate Histopathology). Extensive experiments demonstrate that our Risk-Calibrated Loss consistently reduces the Critical Error Rate (CER) for all four datasets, achieving relative safety improvements ranging from 20.0\% (on breast histopathology) to 92.4\% (on prostate histopathology) compared to state-of-the-art baselines such as Focal Loss. These results confirm that our method offers a superior safety-accuracy trade-off across both CNN and Transformer architectures.
\end{abstract}

\begin{IEEEkeywords}
Medical Image Classification, Risk-Calibrated Learning, Trustworthy AI, Safety-Critical AI.
\end{IEEEkeywords}

\section{Introduction}
\label{sec:introduction}

In high-stakes medical diagnosis, the cost of an error is rarely symmetric. Although deep learning models have achieved expert-level performance in domains ranging from dermatology \cite{esteva2017dermatologist} to histopathology \cite{litjens2017survey}, they are typically optimized using success objectives that treat every misclassification as an equivalent failure (that is, all errors have the same harm). This equivalence is dangerous in clinical practice. 
Confusing two benign skin lesions is a manageable disagreement; but confusing a malignant melanoma with a benign nevus is a potential fatality. Despite this disparity, standard loss functions, including cross-entropy and its variants, often fail to distinguish between safe confusions and fatal diagnostic misses.

Current solutions to this problem generally fall into two categories: addressing data imbalance (e.g., Weighted Cross-Entropy\cite{shannon1948mathematical}) or focusing on hard samples (e.g., Focal Loss\cite{lin2017focal}). 
While these methods effectively prevent the model from ignoring rare classes or easy backgrounds, they remain semantically blind to the direction and harm level of the error. 
A weighted loss may penalize a false negative heavily simply because the positive class is rare, not because it is dangerous. 
Similarly, techniques like Label Smoothing \cite{szegedy2016rethinking} improve model calibration but do not explicitly forbid the specific inter-class confusions that clinicians fear most. 
Consequently, models often achieve high aggregate F1-scores while silently accumulating an unacceptable rate of critical false negatives, effectively trading patient safety for statistical performance. Recent efforts to improve model safety have explored post-hoc targeted error correction, utilizing secondary classifiers to distinguish between acceptable human-like mistakes and severe structural errors \cite{mohammadiseif2026icpr}. While effective, these pipeline-based corrections introduce additional computational overhead during inference. To resolve this alignment problem without requiring secondary models, we propose a Risk-Calibrated Loss technique.

Instead of treating classes as flat, independent entities, we embed the clinical hierarchy (e.g., Benign vs. Malignant) directly into the loss landscape via a confusion-aware severity matrix $\mathbf{M}$. 
By applying asymmetric penalties to structurally distinct errors, our method forces the optimization process to prioritize the elimination of Fatal errors (Malignant $\to$ Benign) over safe, intra-class disagreements. 
Unlike hierarchical methods that enforce symmetric distance penalties \cite{bertinetto2020making}, our technique explicitly models the asymmetry inherent in medical decision-making. We demonstrate the universality of this technique in four different medical imaging benchmarks: 
(1) Brain Tumor MRI (Radiology), 
(2) ISIC 2018 (Dermoscopy - skin lesions), 
(3) BreaKHis (Breast Histopathology), and 
(4) SICAPv2 (Prostate Histopathology). 
Our experiments show that this simple plug-and-play objective reduces Critical Error Rates (CER) by up to 92.4\% on prostate histopathology and 53.7\% on skin lesion benchmarks compared to strong baselines, including Focal Loss and Label Smoothing, without requiring complex architectural changes.

The main contributions of this paper are:
\begin{itemize}
    \item \textbf{Taxonomy of Error:} We propose a formal categorization of medical AI errors into Visual Ambiguity (safe intra-class errors) and Semantic Incoherence (Type I and Type II failures), where Type II is penalized more, given that it may imply a fatality, providing a new lens for evaluating model safety beyond simple accuracy.
    
    \item \textbf{Method:} We introduce Risk-Calibrated Loss (RCL), a parameter-efficient objective ($\alpha, \beta$) that translates hierarchical medical knowledge into a dense cost matrix, strictly suppressing fatal Type II errors.
    
    \item \textbf{Validation:} We demonstrate the efficacy of our approach across four diverse imaging modalities (MRI, Dermoscopy, and two Histopathology datasets), showing that RCL reduces Critical Error Rates (CER) by large margins (up to 92\%) vs. standard cost-sensitive baselines.

    \item \textbf{Analysis:} We provide an empirical analysis of the Safety-Accuracy Pareto Frontier, showing that our method provides a superior trade-off between diagnostic precision and clinical safety than state-of-the-art approaches.
\end{itemize}

The paper is organized as follows. Section~\ref{sec:related_work} covers related work. Our methodology is in Section~\ref{sec:methodology}, while Section~\ref{sec:experimental_setup} gives the experimental setup. In Section~\ref{sec:results} we present results and discuss them, ending in Section~\ref{sec:conclusions} with conclusions.

\section{Related Work}
\label{sec:related_work}
\subsection{Deep Learning in Medical Imaging}
Deep convolutional neural networks (CNNs) have become the standard for medical image analysis, transitioning from handcrafted features to end-to-end learning paradigms \cite{litjens2017survey}. Landmark studies have demonstrated that CNNs and other deep learning architectures can achieve expert-level performance in complex clinical tasks, ranging from classifying skin lesions \cite{esteva2017dermatologist} to grading histological specimens. However, while these models frequently match or exceed human accuracy, they are traditionally optimized using scalar metrics (e.g., accuracy, AUC) that do not account for the varying consequences of different error types. As noted by Freitas et al. \cite{freitas2009cost}, medical decision-making is inherently cost-sensitive; a false negative (missed diagnosis) often carries fatal consequences compared to a false positive. Consequently, optimizing for average performance without integrating expert knowledge regarding error severity remains a critical limitation in clinical deployment \cite{srivastava2023severity}.

\subsection{Addressing Imbalance and Sample Hardness}
To mitigate the risks associated with data bias, various cost-sensitive and re-weighting strategies have been proposed. 
Khan et al. \cite{khan2017cost} introduced a cost-sensitive deep learning technique that jointly optimizes class-dependent costs and network parameters to address class imbalance without altering the data distribution. 
Similarly, to address the issue of hard samples, cases where the model has low confidence, Lin et al. \cite{lin2017focal} proposed Focal Loss which forces the model to focus on sparse, hard-to-classify examples. 
While effective at improving general discrimination, these methods primarily address frequency (minority classes) or statistical hardness (low confidence). 
They remain semantically blind to the direction of the error; for instance, Focal Loss penalizes a hard Benign $\to$ Malignant error identically to a hard Malignant $\to$ Benign error, failing to capture the asymmetric safety risk. Furthermore, building robust models requires addressing intrinsic data complexity and visual hardness \cite{mohammadiseif2026cai}, as well as designing distribution-aware loss functions that go beyond simple mean optimization \cite{mohammadiseif2026bimodalloss}.

\subsection{Hierarchical and Safety-Aware Classification}
Recent approaches have attempted to incorporate semantic structure into the learning process. 
Szegedy et al. \cite{szegedy2016rethinking} introduced Label Smoothing, which improves model calibration and generalization by discouraging over-confident predictions, though it applies a uniform penalty across all classes. 
More explicitly, Bertinetto et al. \cite{bertinetto2020making} proposed Hierarchical Cross-Entropy (HXE) to leverage taxonomic trees, ensuring that mistakes remain semantically close to the ground truth (e.g., confusing two dog breeds is preferred over confusing a dog with a car). 
In the medical domain, Yu et al. \cite{yu2025hierarchical} recently extended this concept using prototypical decision trees to prioritize less severe misclassifications in the analysis of skin lesions.

However, a fundamental limitation of HXE and similar distance-based methods \cite{bertinetto2020making,srivastava2023severity} is their assumption of error symmetry. 
These techniques typically penalize errors based on graph distance, treating a prediction of Class A given Class B effectively as Class B given Class A. 
In safety-critical medical diagnostics, errors are fundamentally asymmetric: distinguishing Malignant from Benign is a safety requirement, whereas distinguishing between Benign subtypes is often merely a taxonomic preference. 

\subsection{Cost-Sensitive Learning and Error Taxonomy}
Standard approaches to balancing False Positives and False Negatives are based on Cost-Sensitive Learning (CSL) \cite{elkan2001foundations}, which assigns static penalties to specific error types. Although CSL effectively shifts the decision boundary to minimize aggregate risk, it typically treats all inter-class confusions within a category as equally severe, often weighting penalties based purely on class frequency or flat, arbitrary costs rather than structural relationships. However, in medical taxonomy, errors exhibit different levels of \textit{semantic incoherence}. As defined later, these range from plausible visual ambiguities to catastrophic failures. Recent work in trustworthy AI \cite{amodei2016concrete} suggests that safety requires distinguishing between these fine-grained ambiguities and high-stakes failures, yet standard loss functions rarely model this hierarchy explicitly during optimization. 

Our work bridges this gap by introducing a risk-calibrated objective that explicitly models this asymmetry. Unlike standard cost matrices that apply global penalties, our technique decouples the suppression of fatal errors from the learning of fine-grained features, achieving safety without requiring complex architectural changes. A conceptual comparison of how our approach contrasts with standard loss functions is summarized in Table \ref{tab:loss_comparison}.

\begin{table}[h]
\centering
\caption{Comparison of Baseline Loss Functions vs. RCL.}
\label{tab:loss_comparison}
\resizebox{\columnwidth}{!}{%
\begin{tabular}{l|c|c|l}
\hline
\textbf{Method} & \textbf{Optimization Focus} & \textbf{Penalty Type} & \textbf{Safety-Aware} \\ \hline
Standard CE \cite{shannon1948mathematical} & Accuracy & Uniform & No \\
Focal Loss \cite{lin2017focal} & Hard Samples & Dynamic (Confidence) & No \\
HXE \cite{bertinetto2020making} & Taxonomy & Symmetric Distance & Partially \\
\textbf{Ours (RCL)} & Clinical Risk & Asymmetric (Semantic) & \textbf{Yes} \\ \hline
\end{tabular}%
}
\end{table}

\section{Methodology}
\label{sec:methodology}

\subsection{Problem Formulation: Visual vs. Semantic Errors}
\label{sec:problem_formulation}

Machine learning (ML) models have achieved remarkable success in computer vision, often surpassing human performance in specific tasks. However, despite their high accuracy, these models remain prone to failures that differ fundamentally from human errors. When a human misclassifies an image, the mistake usually stems from visual ambiguity or fine-grained similarity (e.g., confusing a Siberian Husky with a Malamute). In contrast, deep neural networks often produce high-confidence predictions that are semantically incoherent to humans (e.g., confusing a clearly Benign mole with a Malignant melanoma). These categorical failures are not merely accuracy penalties; they erode trust and pose severe risks in safety-critical domains \cite{omrani2022trust,raji2023concrete,ryan2020ai}. Analogous to autonomous driving, where fatal accidents often stem from catastrophic perception failures (such as failing to distinguish a white truck against a bright sky), medical AI must be evaluated not just on how often it is right, but on how plausible its mistakes are.

\textbf{Definition (Semantic Incoherence):} A prediction is semantically incoherent when the predicted class $\hat{y}$ and true class $y$ belong to different clinical superclasses (e.g., predicting a clearly Benign mole as a Malignant melanoma), representing a structural failure rather than a fine-grained visual confusion.

\begin{figure}[t]
\centering
\includegraphics[width=0.49\columnwidth]{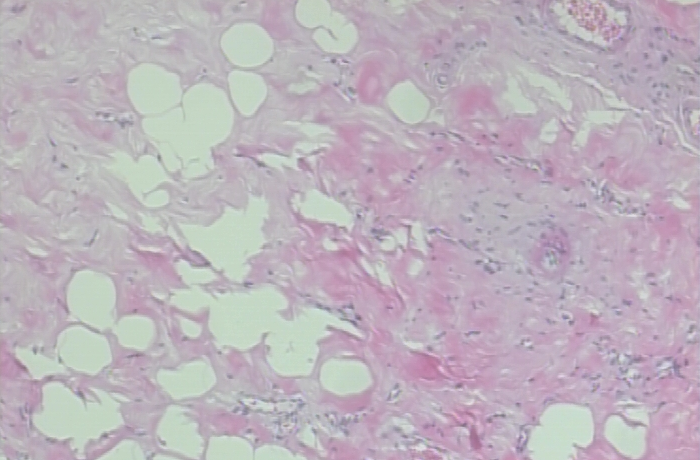} \hfill
\includegraphics[width=0.49\columnwidth]{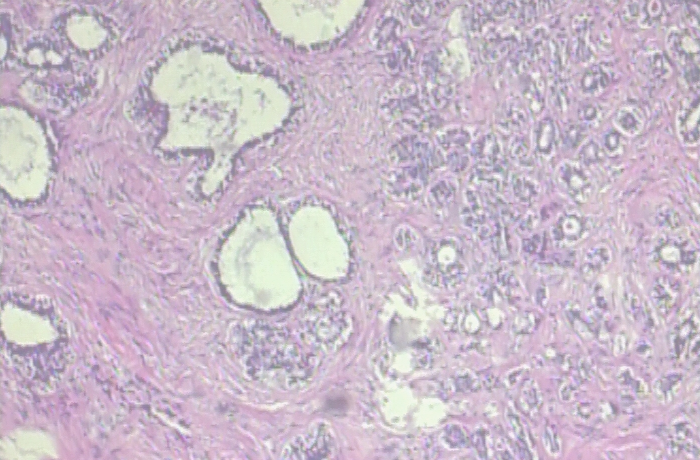} \\
\vspace{2pt}
\scriptsize{(a) \textbf{Visual Ambiguity:} Ambiguity between two Benign subtypes.} \\
\vspace{8pt}

\includegraphics[width=0.49\columnwidth]{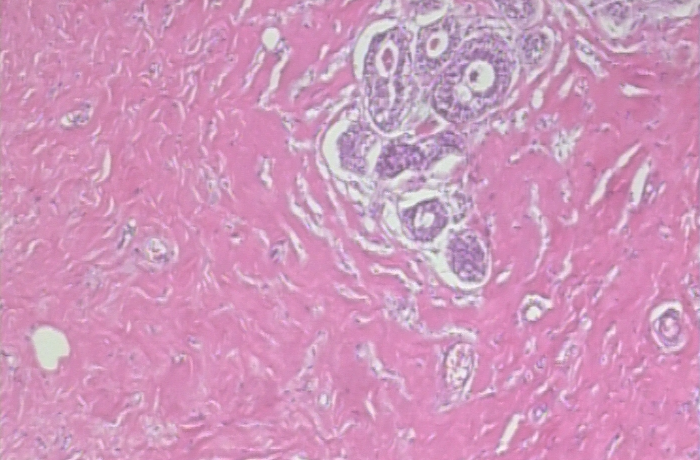} \hfill
\includegraphics[width=0.49\columnwidth]{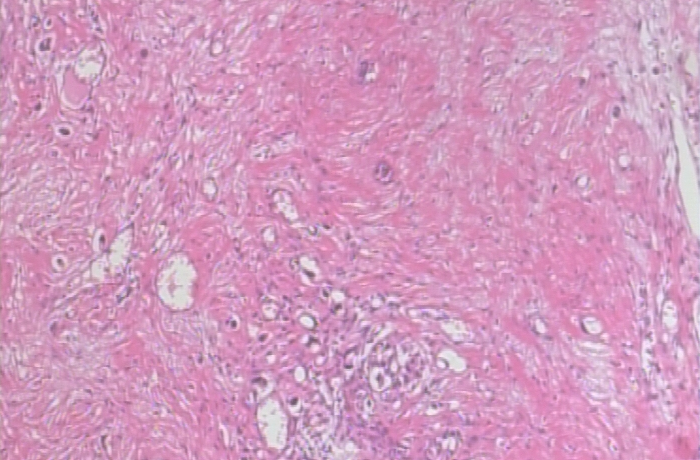} \\
\vspace{2pt}
\scriptsize{(b) \textbf{Type I (False Positive):} Benign (Left) confused with Malignant (Right).} \\
\vspace{8pt}

\includegraphics[width=0.49\columnwidth]{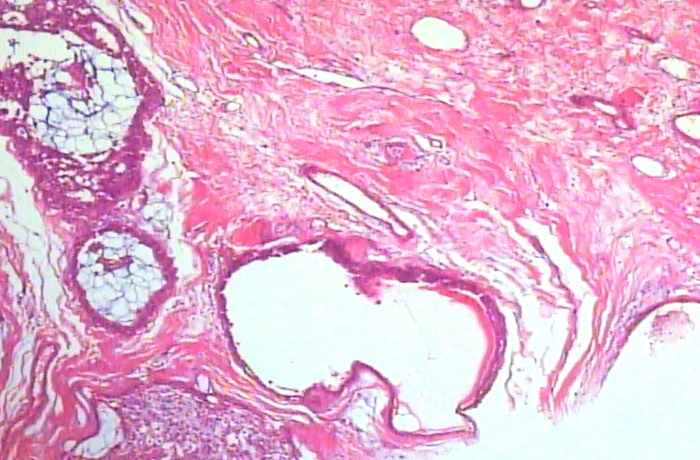} \hfill
\includegraphics[width=0.49\columnwidth]{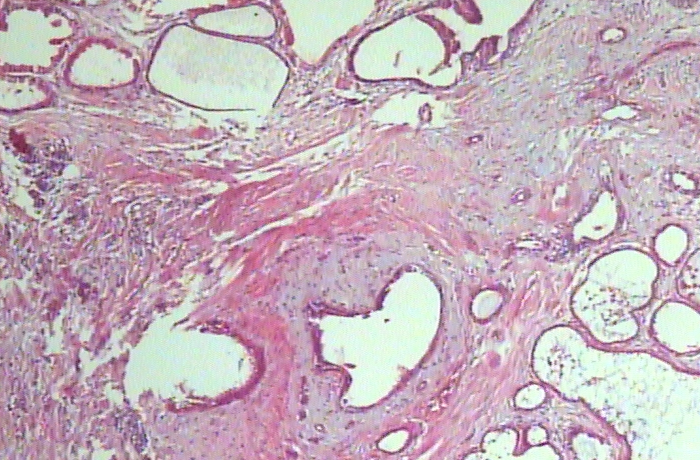} \\
\vspace{2pt}
\scriptsize{(c) \textbf{Type II (Fatal/False Negative):} Malignant (Left) misclassified as Benign (Right).} \\
\vspace{4pt}

\caption{\textbf{The Spectrum of Errors (BreaKHis dataset).} 
(a) \textbf{Visual Ambiguity:} Confusing visual lookalikes (e.g., Adenosis vs. Fibroadenoma) is acceptable. 
(b) \textbf{Type I (Costly):} A False Alarm where benign tissue is flagged as cancer. 
(c) \textbf{Type II (Fatal):} A catastrophic failure where an obvious Mucinous Carcinoma is classified as a Benign Adenoma.
}
\label{fig:error_spectrum}
\vspace*{-5mm}
\end{figure}

Medical diagnostics requires a strict taxonomy based on patient outcomes. We therefore categorize errors into three distinct levels of severity, visually summarized in Figure \ref{fig:error_spectrum}:

\begin{enumerate}
    \item \textbf{Visual Ambiguity (Intra-Class Error):} The predicted class $\hat{y}$ differs from the ground truth $y$, but both belong to the same superclass (e.g., $\hat{y}, y \in \text{Benign}$). This represents a safe disagreement arising from fine-grained visual similarity, analogous to human inter-rater variability.
    
    \item \textbf{Type I Error (False Positive):} A form of semantic incoherence where the prediction crosses the superclass boundary: the model predicts Malignant when the ground truth is Benign. While clinically fail-safe (triggering verification), it imposes severe psychological and financial costs.
    
    \item \textbf{Type II Error (Fatal or False Negatives):} The most severe form of semantic incoherence (Inter-Class Error). The model predicts Benign when the ground truth is Malignant. Unlike visual ambiguity, this structural failure results in untreated disease progression and is strictly unacceptable in clinical deployment.
\end{enumerate}

Our goal is to train a deep neural network that minimizes the total loss while strictly prioritizing the elimination of Critical Type II Errors, acknowledging that while Type I errors are costly, Type II errors are irreversible.

\subsection{Risk-Calibrated Loss}
To embed these safety constraints directly into the optimization process, we propose the Risk-Calibrated Loss ($\mathcal{L}_{RC}$). We define a clinical severity matrix $\mathbf{M} \in \mathbb{R}^{K \times K}$, where entry $M_{y, k}$ represents the penalty multiplier for misclassifying true class $y$ as class $k$. The matrix is constructed according to the hierarchy defined above: we assign $M_{visual}=1$ for safe fine-grained errors, $M_{type1}=\alpha$ for false positives, and $M_{fatal}=\beta$ for critical false negatives. 

The Risk-Calibrated Loss is formally defined as:
\begin{equation}
\mathcal{L}_{RC}(x, y) = M_{y, \hat{y}} \cdot \mathcal{L}_{CE}(x, y)
\end{equation}
where $\mathcal{L}_{CE}$ is the standard Cross-Entropy loss \cite{shannon1948mathematical}, and $\hat{y}$ is the current prediction of the model. Hyperparameters $\alpha \ge 1$ and $\beta \ge \alpha$ define Type I Penalty and Type II Penalty to control superclass confusion and fatal errors, respectively. By dynamically scaling the loss magnitude based on the current prediction's severity, the gradient signal for critical errors is amplified by a factor $\beta$, forcing the optimizer to escape regions of the parameter space prone to fatal misclassifications.

To illustrate, consider a dataset with two Benign subtypes ($B_1, B_2$) and two Malignant subtypes ($M_1, M_2$). Let $\hat{B}$ and $\hat{M}$ denote their corresponding model predictions. The severity matrix $\mathbf{M}$ takes the following block structure, explicitly decoupling safe intra-class confusions from structural semantic errors:

\begin{equation}
\mathbf{M} = 
\begin{array}{c@{\!}c}
&\begin{array}{cccc} \hat{B}_1 & \hat{B}_2 & \hat{M}_1 & \hat{M}_2 \end{array} \\[1ex]
\begin{array}{r}
B_1 \\
B_2 \\
M_1 \\
M_2 
\end{array} &
\left[ \begin{array}{cccc}
1 & 1 & \alpha & \alpha \\
1 & 1 & \alpha & \alpha \\
\beta & \beta & 1 & 1 \\
\beta & \beta & 1 & 1 
\end{array} \right]
\end{array}
\label{eq:matrix}
\end{equation}

Here, the top-right quadrant ($\alpha$) penalizes Type I False Positives, while the bottom-left quadrant applies the severe Type II Penalty ($\beta$) for fatal misses. The diagonal blocks remain as $1$, treating fine-grained visual ambiguity identically to standard cross-entropy.

\subsection{Contrast with Standard Objectives}
It is important to distinguish our Risk-Calibrated Loss from standard Weighted Cross-Entropy (WCE). WCE assigns a static weight $w_y$ to each class $y$, typically inversely proportional to the class frequency ($w_y \propto 1/N_y$), mitigating data imbalance. Mathematically, the gradient of WCE for a sample of class $y$ is scaled by $w_y$ regardless of whether the model predicted a safe neighbor or a fatal opposite. 
However, WCE is agnostic to the nature of the misclassifications; it penalizes a Benign $\to$ Benign error identically to a Malignant $\to$ Benign error if class frequencies are similar. In contrast, our technique employs a dynamic, confusion-dependent cost matrix that scales the gradient based on the semantic severity of the prediction ($M_{y, \hat{y}}$). 
This creates an anisotropic optimization landscape where the uphill cost of making a fatal error is significantly steeper than that of a safe error, effectively embedding clinical safety constraints directly into the model's learning trajectory.

\section{Experimental Setup}
\label{sec:experimental_setup}

To validate the generality of our Risk-Calibrated Loss, we conduct experiments across four medical imaging datasets spanning varying modalities (MRI, Dermoscopy, Histopathology) and magnifications.

\subsection{Datasets and Hierarchies}

We employ a consistent hierarchical protocol across all datasets: classes are mapped to a binary superclass structure where $S=0$ denotes ``Benign/Non-Critical'' and $S=1$ denotes ``Malignant/Critical.'' The specific configurations are detailed below and summarized in Table \ref{tab:datasets}.

\begin{enumerate}
    \item \textbf{Brain Tumor MRI (Radiology):} We utilize the dataset curated by Nickparvar \cite{msoud_nickparvar_2026}, which aggregates 7,023 MRI images from multiple open-source repositories (including Figshare, SARTAJ, and Br35H). It contains four classes: Glioma, Meningioma, Pituitary tumors, and No Tumor. We define ``No Tumor'' as Benign (S=0) and all tumor types as Critical (S=1).
    \item \textbf{ISIC 2018 (Dermoscopy):} A 7-class skin lesion dataset \cite{codella2019skin}. We group Melanoma (MEL), Basal Cell Carcinoma (BCC), and Actinic Keratosis (AKIEC) as Malignant ($S=1$). The remaining classes (NV, BKL, DF, VASC) are grouped as Benign ($S=0$).
    \item \textbf{BreaKHis (Breast Histopathology):} Contains microscopic images of breast tumor tissue. The dataset includes four benign subtypes (Adenosis, Fibroadenoma, Phyllodes Tumor, Tubular Adenoma) and four malignant subtypes (Ductal Carcinoma, Lobular Carcinoma, Mucinous Carcinoma, Papillary Carcinoma) \cite{ambarish2019breakhis}.
    \item \textbf{SICAPv2 (Prostate Histopathology):} A patch-based dataset for prostate cancer classification \cite{silva2020going}. We map Non-Cancerous (NC) tissue to Benign ($S=0$). Gleason Grades 3, 4, and 5, indicating an increase in cancer severity, are all assigned to the Malignant superclass ($S=1$).
\end{enumerate}

\begin{table}[t]
\centering
\caption{dataset Statistics and Safety Hierarchies\newline
($S=0$: Benign, $S=1$: Malignant).}
\label{tab:datasets}
\resizebox{\columnwidth}{!}{%
\begin{tabular}{l|c|c|l}
\hline
\textbf{dataset} & \textbf{Modality} & \textbf{Classes} & \textbf{Critical Split ($S=0$ vs. $S=1$)} \\ \hline
Brain Tumor & MRI & 4 & No Tumor vs. [Glioma, Menin., Pituitary] \\
ISIC 2018 & Dermoscopy & 7 & [NV, BKL, DF, VASC] vs. [MEL, BCC, AKIEC] \\
BreaKHis & Histo (Breast) & 8 & 4 Benign Subtypes vs. 4 Malignant Subtypes \\
SICAPv2 & Histo (Prostate) & 4 & NC vs. [Grade 3, Grade 4, Grade 5] \\ \hline
\end{tabular}%
}
\end{table}

\subsection{Implementation Details}

\textbf{Architectures:} We evaluate two distinct backbone architectures to analyze architectural sensitivity: ResNet-50 \cite{he2016deep}, representing standard Convolutional Neural Networks (CNNs), and ViT-B16 \cite{dosovitskiy2020image}, representing Vision Transformers. Both models are initialized with ImageNet pre-trained weights.

\textbf{Training Configuration:} All models are trained using the AdamW optimizer with a learning rate of $10^{-4}$ and a batch size of 32. We employ a cosine annealing learning rate scheduler. Images are resized to $224 \times 224$ pixels with standard augmentations (random rotation, flip, and normalization).

\textbf{Baselines:} We compare our method against four state-of-the-art baselines:
\begin{itemize}
    \item \textbf{Cross-Entropy (CE) \cite{shannon1948mathematical}:} Standard baseline.
    \item \textbf{Weighted Cross-Entropy (WCE) \cite{shannon1948mathematical}:} Weights are set inversely proportional to class frequency ($w_c \propto 1/N_c$).
    \item \textbf{Focal Loss (FL) \cite{lin2017focal}:} Configured with focusing parameters $\gamma=2.0$ and $\alpha=1.0$ to target hard samples.
    \item \textbf{Label Smoothing (LS) \cite{szegedy2016rethinking}:} Smoothing factor set to $\epsilon=0.1$ to mitigate overconfidence.
\end{itemize}

\textbf{Risk-Calibrated Ablation Study:} To determine the optimal balance between superclass consistency ($\alpha$) and the Type II Penalty ($\beta$), we evaluated distinct configurations of our Risk-Calibrated Loss.

\subsection{Evaluation Metrics}
To assess model performance, we utilize three key metrics:
\noindent \textbf{Critical Error Rate (CER):} The primary safety metric, defined as the proportion of Malignant samples incorrectly classified as Benign (False Negatives at the superclass level).
\begin{equation}
    \text{CER} = N_{\text{Fatal}} / N_{\text{Malignant}} \times 100
\end{equation}
where $N_{\text{Fatal}}$ is the count of fatal Type II errors. A lower CER indicates a safer model.\\
\noindent \textbf{F1-Macro:} The harmonic mean of precision ($P_k$) and recall ($R_k$) averaged across all $K$ fine-grained classes. This ensures that improvements in safety do not come at the cost of collapsing performance in rare classes.\\
\noindent \textbf{Standard Accuracy:} The fine-grained classification accuracy, measuring global model correctness across all classes.

\section{Results and Discussion}
\label{sec:results}

In this section, we first report the gains in Critical Error Rate (CER), followed by an explicit analysis of the safety-accuracy trade-off.
\begin{table*}[t]
\caption{Comparative Performance of Risk-Calibrated Loss (RCL) vs. State-of-the-Art Baselines. We report the \textbf{Proposed} configuration ($\alpha=5, \beta=20$) as RCL. \textbf{CER} denotes Critical Error Rate (Lower is Better). \textbf{Acc} denotes Standard Fine-Grained Accuracy. Best Safety results are in \textbf{bold}.}
\label{tab:main_results}
\centering
\begin{tabular}{l|l||ccc||ccc||c}
\hline
\textbf{Dataset} & \textbf{Model} & \multicolumn{3}{c||}{\textbf{Baseline: Focal Loss} \cite{lin2017focal}} & \multicolumn{3}{c||}{\textbf{Ours: RCL (Proposed)}} & \textbf{CER Improvement} \\
 & & CER ($\downarrow$) & F1-Macro & Acc & \textbf{CER} ($\downarrow$) & F1-Macro & Acc & Abs. (Rel. \% $\downarrow$) \\
\hline
\hline
\textbf{Brain MRI} & ResNet-50 & 0.00\% & 0.995 & 99.5\% & \textbf{0.00\%} & 0.993 & 99.3\% & 0.00\% (-) \\
(Radiology) & ViT-B16 & 0.20\% & 0.990 & 99.0\% & \textbf{0.10\%} & 0.985 & 98.6\% & -0.10\% (-50.0\%) \\
\hline
\textbf{BreaKHis} & ResNet-50 & 2.95\% & 0.174 & 53.8\% & \textbf{2.36\%} & 0.278 & 56.0\% & -0.59\% (-20.0\%) \\
(Breast Histo) & ViT-B16 & 0.89\% & 0.259 & 54.8\% & \textbf{0.59\%} & 0.101 & 47.8\% & -0.30\% (-33.3\%) \\
\hline
\textbf{SICAPv2} & ResNet-50 & 9.07\% & 0.561 & 63.6\% & \textbf{6.23\%} & 0.599 & 69.3\% & -2.84\% (-31.3\%) \\
(Prostate Histo) & ViT-B16 & 10.69\% & 0.666 & 71.1\% & \textbf{0.81\%} & 0.436 & 57.8\% & -9.88\% (-92.4\%) \\
\hline
\textbf{ISIC 2018} & ResNet-50 & 39.41\% & 0.692 & 80.9\% & \textbf{18.24\%} & 0.595 & 76.1\% & -21.17\% (-53.7\%) \\
(Dermoscopy) & ViT-B16 & 43.32\% & 0.590 & 76.3\% & \textbf{24.10\%} & 0.541 & 69.0\% & -19.22\% (-44.4\%) \\
\hline
\end{tabular}
\end{table*}

\subsection{Performance Evaluation on Fatal Errors}
The primary objective of this work was to minimize Fatal Type II Errors. We benchmark our Risk-Calibrated Learning (RCL) technique using the \textit{Proposed} configuration ($\alpha=5, \beta=20$) against the industry-standard Focal Loss \cite{lin2017focal}. We select Focal Loss as the representative baseline in Table \ref{tab:main_results} due to its prevalence in recent literature, though we analyze the full comparative landscape including Weighted Cross-Entropy (WCE) and Label Smoothing (LS) below. Table \ref{tab:main_results} demonstrates that RCL consistently reduces the Critical Error Rate (CER) across all four diverse use cases.

\subsubsection{Comparison with Cost-Sensitive Baselines}
While Focal Loss targets hard samples, we also evaluated Weighted Cross-Entropy (WCE) and Label Smoothing (LS). Consistent with cost-sensitive literature, WCE improved safety (reducing CER to 25.73\% on ISIC 2018) compared to standard Cross-Entropy (39.09\%), but at the cost of significantly higher variance and lower precision. Label Smoothing improved model calibration but failed to reduce fatal errors (CER 38.44\%), performing nearly identically to the unweighted baseline. RCL outperformed all strategies, achieving a CER of 18.24\%, a 29\% relative improvement over WCE, demonstrating that dynamic, confusion-aware penalties are superior to static class-weighting.

\subsubsection{High-Difficulty Scenarios (Dermoscopy)}
The most significant impact is observed in the ISIC 2018 skin lesion dataset, known for high class imbalance and visual ambiguity. Standard baselines struggle here as Focal Loss yields a CER of 39.41\% on ResNet-50, implying that nearly 40\% of malignant cases were missed. By embedding the cost matrix into the loss function, RCL reduces this to 18.24\%, a relative safety improvement of 53.7\%. This indicates that semantic penalties are more effective than frequency-based re-weighting when the decision boundary is subtle.

\subsubsection{Architectural Comparison}
A key observation is the differential impact of RCL on Convolutional Neural Networks (ResNet-50) versus Vision Transformers (ViT-B16). While both architectures benefit from the risk-calibrated penalty, ViT models often show larger relative gains in safety.
For instance, on the SICAPv2 prostate cancer dataset, the ResNet-50 model improved its safety metric by 31.3\%, whereas the ViT-B16 model achieved a 92.4\% relative reduction in fatal errors (dropping CER from 10.69\% to 0.81\%). 

This suggests that the global attention mechanism of Transformers may be more adaptable to semantic constraints.
Unlike the local receptive fields of CNNs, which primarily capture local textural features and can easily be fooled by spurious local correlations (e.g., confusing a dark benign patch for melanoma), the global context captured by ViTs allows the network to integrate information from the entire lesion structure. Consequently, when penalized by the RCL objective, the ViT optimization process can shift decision boundaries more effectively, learning holistic structural representations that avoid fatal regions of the latent space. 

\subsection{The Safety-Accuracy Trade-off}

\begin{figure}[t]
\centerline{\includegraphics[width=\columnwidth]{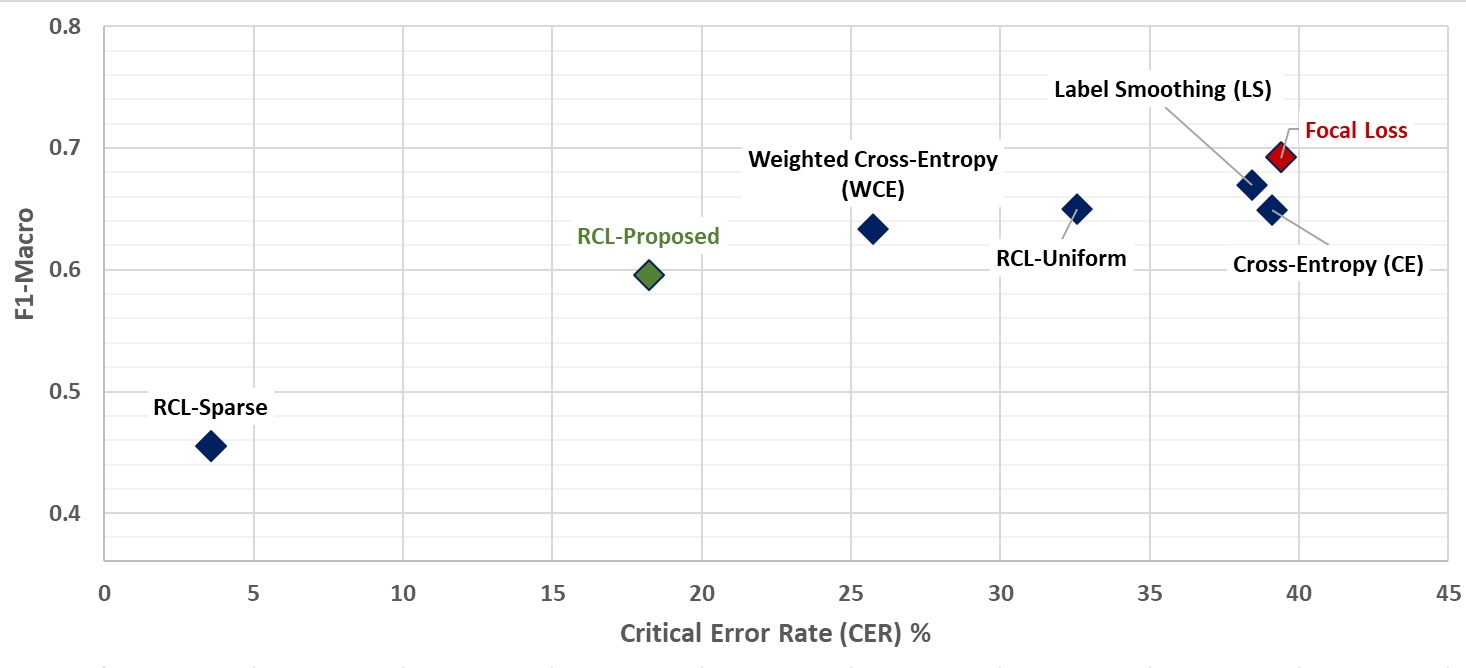}} 
\caption{\textbf{Safety vs. Accuracy Trade-off (ISIC 2018, ResNet-50).} The scatter plot compares RCL against standard baselines (CE, WCE, Focal, LS). The X-axis represents the overall F1-Macro score (higher is better), while the Y-axis represents the Critical Error Rate (CER, lower is safer). Each data point corresponds to a specific loss function's performance. The Focal Loss baseline (Red) remains in the high-risk, high-accuracy zone, whereas the RCL-Proposed method (Green) explicitly shifts the model to the safe Top-Left corner, prioritizing the reduction of fatal errors over marginal gains in fine-grained accuracy.}
\label{fig:tradeoff}
\vspace*{-2mm}
\end{figure}

Our results reveal a dual dynamic. In structured tasks like Histopathology (BreaKHis, SICAPv2), penalizing fatal errors actually improves standard accuracy (e.g., RCL boosts ResNet-50 accuracy on BreaKHis from 53.8\% to 56.0\% compared to Focal Loss). This suggests that for distinct pathologies, safety constraints help sharpen the decision boundary.
However, in highly ambiguous domains like Dermoscopy (ISIC 2018), a necessary trade-off emerges (Figure \ref{fig:tradeoff}).

While Focal Loss achieves a higher F1-Macro score (0.692) than RCL (0.595), deeper inspection shows that this performance is driven by correctly identifying the majority benign classes while permitting a high rate of fatal errors. In contrast, RCL accepts a reduction in standard accuracy (80.9\% $\to$ 76.1\%) to secure a major gain in sensitivity to malignancy, raising the detection rate of critical cases from 60.6\% to 81.8\% (as implied by the reduction in CER). In a clinical support context, this trade-off is often preferred: a model that flags 81\% of cancers with some false positives is clinically actionable, whereas a model that misses 40\% of cancers is unsafe.

\begin{table}[t]
\centering
\caption{Hyperparameter Configurations for Ablation Study. $\alpha$ defines the Type I Penalty (False Positive); $\beta$ defines the Type II Penalty (Fatal).}
\label{tab:ablation_configs}
\resizebox{\columnwidth}{!}{%
\begin{tabular}{l|cc|l}
\hline
\textbf{Configuration} & \textbf{$\alpha$} & \textbf{$\beta$} & \textbf{Motivation / Hypothesis} \\ \hline
RCL-Light & 5.0 & 5.0 & Baseline check for stability (accuracy) \\
RCL-Balanced & 2.0 & 10.0 & Attempt to recover fine-grained accuracy \\
RCL-StructSafe & 5.0 & 10.0 & Stronger structure, milder safety constraint \\
RCL-Uniform & 10.0 & 10.0 & Coarse, symmetric high penalty \\
RCL-Sparse & 1.0 & 20.0 & Targeted solely on Critical Errors \\
RCL-Proposed & 5.0 & 20.0 & Optimal trade-off (Structure + Safety) \\
RCL-HighStruct & 10.0 & 20.0 & Max regularization (Fix Semantic Incoherence) \\ \hline
\end{tabular}%
}
\end{table}

\subsection{Ablation Study: The Necessity of Asymmetric Penalties}

To validate our Risk-Calibrated Loss, we conducted an ablation study to isolate the impact of the Type I Penalty ($\alpha$) and the Type II Penalty ($\beta$). For this, we evaluate seven configurations of the Risk-Calibrated Loss, detailed in Table \ref{tab:ablation_configs}. These range from \textit{Light} penalties (to check stability) to \textit{High Structure} settings (to enforce strict hierarchy). We also analyze two datasets to illustrate different failure modes.

For a new, unseen dataset, clinicians or engineers should tune these parameters by first defining the Type II Penalty ($\beta$) to the clinical severity of the disease (e.g., mortality risk), and subsequently tuning the Type I Penalty ($\alpha$) based on the maximum tolerable false-positive rate on a validation set to avoid excessive alarm fatigue.

\begin{figure}[t]
\centerline{\includegraphics[width=0.8\columnwidth]{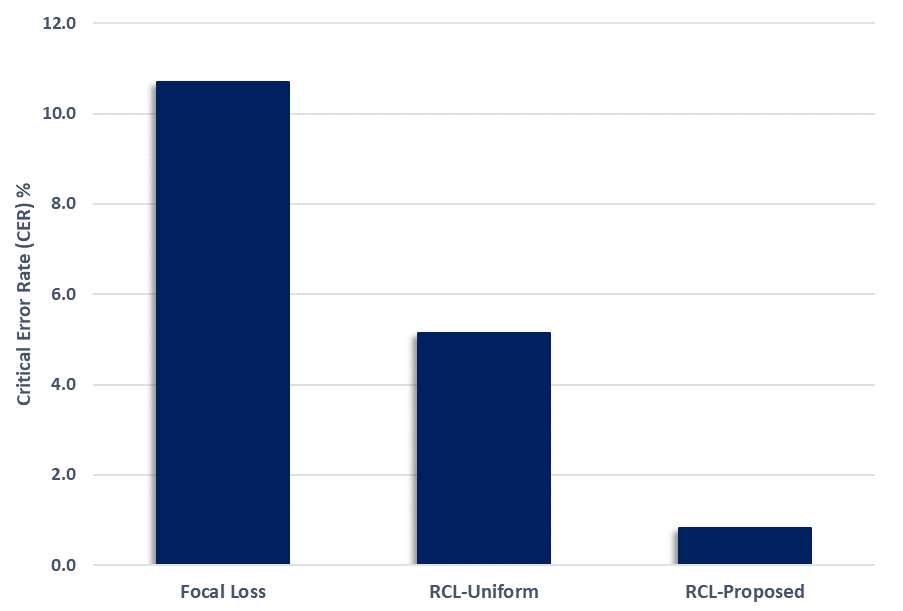}} 
\caption{\textbf{Ablation Study on SICAPv2 (ViT-B16).} The bar chart illustrates the impact of different penalty configurations on model safety. The X-axis categorizes the tested loss configurations, while the Y-axis measures the Critical Error Rate (CER). The \textit{Staircase to Safety} trend demonstrates that while a \textit{Uniform} configuration ($\alpha=10, \beta=10$) reduces some errors relative to the baseline, only the \textit{Proposed} configuration ($\alpha=5, \beta=20$) effectively eliminates fatal errors (achieving a CER $<$ 1\%).}
\label{fig:ablation_bar}
\vspace*{-2mm}
\end{figure}

\subsubsection{Hyperparameter Analysis on ISIC 2018}
Table \ref{tab:ablation} compares three penalization strategies on the difficult ISIC 2018 dataset (ResNet-50).

\begin{table}[t]
\caption{Baseline Comparison and Ablation Study (ISIC 2018, ResNet-50).}
\label{tab:ablation}
\centering
\begin{tabular}{l|cc||c|c}
\hline
\textbf{Method / Configuration} & $\mathbf{\alpha}$ & $\mathbf{\beta}$ & \textbf{CER ($\downarrow$)} & \textbf{F1-Macro ($\uparrow$)} \\
\hline
Standard CE \cite{shannon1948mathematical} & - & - & 39.09\% & 0.649 \\ 
Label Smoothing (LS) \cite{szegedy2016rethinking}& - & - & 38.44\% & 0.669 \\ 
Weighted CE (WCE) \cite{shannon1948mathematical} & - & - & 25.73\% & 0.633 \\ 
Baseline (Focal) \cite{lin2017focal} & - & - & 39.41\% & \textbf{0.692} \\
\hline
RCL-Uniform & 10 & 10 & 32.57\% & 0.650 \\
RCL-Sparse & 1 & 20 & \textbf{3.58\%} & 0.455 \\
\textbf{RCL-Proposed} & \textbf{5} & \textbf{20} & 18.24\% & 0.595 \\
\hline
\end{tabular}
\end{table}

The \textit{Uniform} configuration ($\alpha=10, \beta=10$) yielded a CER of 32.57\%, only a marginal improvement over the baseline. We posit that symmetric heavy penalties create a chaotic optimization landscape, preventing the model from prioritizing the critical class. Conversely, the \textit{Sparse} configuration ($\alpha=1, \beta=20$) achieved the lowest CER (3.58\%) but suffered a collapse in F1-score (0.455), indicating the model learned to predict Malignant for most samples. The \textit{Proposed} configuration strikes the necessary balance, reducing fatal errors by over 50\% while maintaining discriminative power.

\subsubsection{Impact on Convergence (SICAPv2)}
On the SICAPv2 dataset (ViT-B16), shown in Figure \ref{fig:ablation_bar}, we observe a clear hierarchy of efficacy. The Baseline Focal Loss resulted in a high CER of 10.69\%. Introducing a \textit{Uniform} configuration reduced this to 5.14\%, but it was the asymmetric \textit{Proposed} configuration that successfully minimized the error to 0.81\%. This confirms that strictly prioritizing the elimination of fatal Type II errors via asymmetry ($\beta > \alpha$) is essential for breaking the safety ceiling that symmetric cost functions cannot overcome.

\subsection{Analysis of Error Distribution \& Clinical Impact}

Beyond aggregate metrics, we analyze if prioritizing safety degrades fine-grained discrimination. Our data refutes this: on ISIC 2018 (ResNet-50), the count of visual ambiguity errors remained stable (113 vs. 109 with RCL), proving the model retains discriminative power for safe cases.
In contrast, the error distribution shifted strictly along the safety axis: Type II Errors (Fatal) dropped by 54\% (121 $\to$ 56), while Type I Errors increased by 43\% (176 $\to$ 252). This confirms that $\mathcal{L}_{RC}$ acts as a semantic margin enforcer, pushing ambiguous samples into the ``safe'' zone without distorting the feature space.
Clinically, this shift converts the AI from a statistical tool into a high-sensitivity safety net. While the increase in Type I errors implies a modest workload increase and a risk of ``alarm fatigue'' (where clinicians might become desensitized to frequent alerts) for pathologists (verifying flagged benign cases), this operational cost is negligible compared to the irreversible nature of a False Negative. By reducing the Critical Error Rate from $\sim$40\% to $\sim$18\%, RCL aligns with the operational reality of healthcare, where the cost of a second look is merely financial, but the cost of a missed diagnosis is mortality.

\subsection{Limitations and Clinical Considerations}
While RCL provides a robust safety net, it relies on a static clinical taxonomy. In complex pathologies with high inter-rater variability, a rigid binary severity matrix may oversimplify the risk continuum; future work should explore probabilistic matrices to model this uncertainty. 

Additionally, aggressively reducing Type II errors inherently increases Type I errors (false positives), which can trigger unnecessary biopsies and elevate costs. Therefore, RCL is best deployed as a sensitive triage tool for expert verification rather than an autonomous diagnostic agent.

\section{Conclusions}
\label{sec:conclusions}

We introduced Risk-Calibrated Learning (RCL) to align deep learning with the asymmetric costs of medical error. By distinguishing visual ambiguity (safe fine-grained errors) from Type II fatal errors, our hierarchical loss suppresses critical failures. Validation across four datasets (MRI, Dermoscopy, Histopathology) confirms that RCL minimizes Critical Error Rates versus baselines, demonstrating that safety stems from a semantically aware landscape rather than architectural complexity. Ultimately, RCL enforces the diagnostic golden rule: re-examining a healthy patient is acceptable; sending an ill patient home is not.

Although this work focuses on classification, future work will extend RCL to dense prediction tasks like object detection and semantic segmentation, where under-segmentation (or missed bounding boxes) constitutes a fatal error, and explore dynamic weighting for multi-modal fusion of imaging and clinical notes to further enhance reliability.

\section*{Acknowledgment}
This work has been partially supported by MCIN/AEI/10.13039/501100011033 under the Mar\'ia de Maeztu Units of Excellence Program (CEX2021-001195-M).

\bibliographystyle{IEEEtran} 
\bibliography{refs}


\end{document}